\documentclass{article}

\usepackage{arxiv}

\usepackage[utf8]{inputenc} 
\usepackage[T1]{fontenc}    
\usepackage{hyperref}       
\usepackage{url}            
\usepackage{booktabs}       
\usepackage{amsfonts}       
\usepackage{nicefrac}       
\usepackage{microtype}      
\usepackage{lipsum}		
\usepackage{graphicx}
\usepackage[numbers]{natbib}
\usepackage{doi}
\usepackage{multirow}%
\usepackage{amsmath,amssymb,amsfonts}%
\usepackage{amsthm}%
\usepackage{mathrsfs}%
\usepackage[title]{appendix}%
\usepackage{xcolor}%
\usepackage{textcomp}%
\usepackage{manyfoot}%
\usepackage{booktabs}%
\usepackage{algorithm}%
\usepackage{algorithmicx}%
\usepackage{algpseudocode}%
\usepackage{listings}%

\title{Reinforcement Learning to improve delta robot throws for sorting scrap metal}

\author{\hspace{1mm}Arthur~Louette\thanks{Montefiore Institute,  University of Liège, Belgium} \\
	\texttt{arthur.louette@uliege.be} \\
	\And
	Gaspard~Lambrechts\textsuperscript{*} \\
	\texttt{gaspard.lambrechts@uliege.be} \\
         \AND
          Damien Ernst\textsuperscript{*}\thanks{LTCI, Telecom Paris, Institut Polytechnique de Paris, France} \\
	\texttt{dernst@uliege.be} \\
	\And
	  Eric Pirard\thanks{Georesources, Mineral Engineering \& Extractive Metallurgy,
        University of Liège, Belgium}\\
	\texttt{eric.pirard@uliege.be} \\
        \And
	  Godefroid Dislaire\textsuperscript{‡}\\
	\texttt{godefroid.dislaire@uliege.be} \\
}

\renewcommand{\headeright}{}
\renewcommand{\undertitle}{}
\renewcommand{\shorttitle}{Reinforcement Learning to improve delta robot throws for sorting scrap metal}

\hypersetup{
pdftitle={Reinforcement Learning to improve delta robot throws for sorting scrap metal},
pdfsubject={Reinforcement learning, Delta robot, robotic sorting line},
pdfauthor={Arthur~Louette, Gaspard~Lambrecht, Damien~Ernst, Eric~Pirard, Godefroid~Disclaire},
pdfkeywords={Delta robots, Reinforcement Learning, Pick-and-Throw, Pybullet},
}

\begin{document}
\maketitle

\begin{abstract}
This study proposes a novel approach based on reinforcement learning (RL) to enhance the sorting efficiency of scrap metal using delta robots and a Pick-and-Place (PaP) process, widely used in the industry. We use three classical model-free RL algorithms (TD3, SAC and PPO) to reduce the time to sort metal scraps. We learn the release position and speed needed to throw an object in a bin instead of moving to the exact bin location, as with the classical PaP technique. Our contribution is threefold. First, we provide a new simulation environment for learning RL-based Pick-and-Throw (PaT) strategies for parallel grippers. Second, we use RL algorithms for learning this task in this environment resulting in 89.32\% accuracy while speeding up the throughput by 51\% in simulation. Third, we evaluate the performances of RL algorithms and compare them to a PaP and a state-of-the-art PaT method both in simulation and reality, learning only from simulation with domain randomisation and without fine tuning in reality to transfer our policies. This work shows the benefits of RL-based PaT compared to PaP or classical optimization PaT techniques used in the industry. The code is available at \href{https://github.com/LouetteArthur/Pick-and-throw/}{https://github.com/LouetteArthur/Pick-and-throw}.
\end{abstract}


\section{Introduction}
\begin{figure}
    \centering
    \begin{minipage}{0.8\textwidth}
        \centering
        \includegraphics[width=\linewidth]{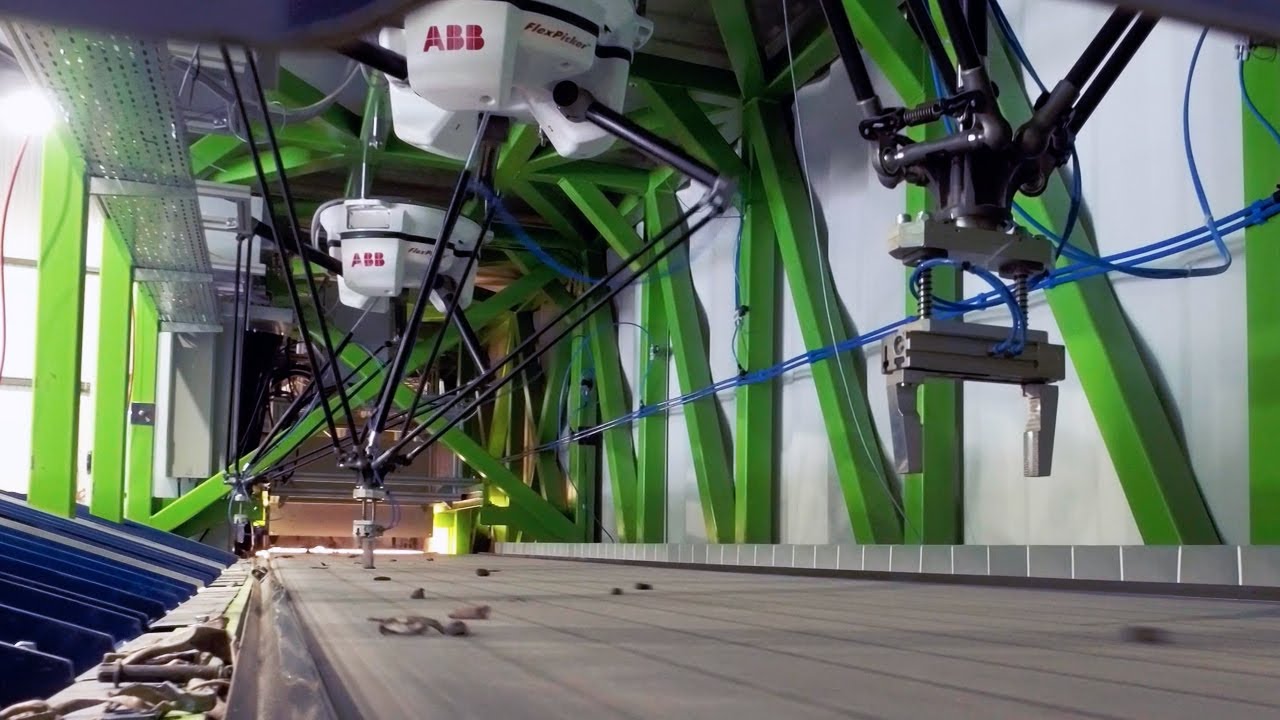}
        \caption{Photo of the PICKIT project at the GeMMe (Georesources, Mineral Engineering \& Extractive Metallurgy) lab with a focus on the ABB IRB 360 Flexpicker robots sorting scrap metal.}
        \label{fig:rob_line}
    \end{minipage}
\end{figure}

The improving performance of contemporary technologies is correlated with an increasing complexity. This complexity presents a significant challenge at the end of life of devices, where the meticulous disassembly of products is essential for recycling and remanufacturing~\citep{Gundupalli2017automatedsortingreview}. This disassembly notably involves segregating individual components and sorting materials into numerous categories. While optical sorting techniques have been in existence for a considerable period of time, it has become evident that there is a pressing requirement to handle a substantially higher volume of throughputs~\citep{raptopoulos_2020}. A novel approach involving a series of inline delta robots has shown promises to solve this issue. More specifically, the GeMMe lab has been developing such a robotic line to efficiently sort scraps of metal on a singular conveyor belt, directing the scraps into multiple designated containers, as illustrated in Figure \ref{fig:rob_line}.

By adopting a Pick-and-Place (PaP) policy, delta robots can sort more than 4000 objects per hour. This approach offers precision and agility, allowing these robots to accurately pick up items from a conveyor belt and precisely place them into designated bins. However, recent research has been looking for a way to speed up this strategy\citep{raptopoulos_2020, chen2019, hassan_2022}. Such new approaches could be to use a Pick-and-Throw (PaT) strategy but it has unfortunately uncovered several challenges. Unlike PaP, where the optimal trajectory is known depending only on the bin, the optimal control for throwing an object while minimizing failed throws and time is not trivial. 

Reinforcement learning (RL) has demonstrated outstanding results in controlling robots from interactions with the environment\citep{zeng2019tossingbot, aractingi2023controlling, Johannink2019residual}. RL can overcome the modelling difficulties that classical control methods face since model-free RL agents learn only from interactions with the environment without having to model its dynamics explicitly. This work considers the problem of finding the release position and the speed of the robot as a one-step RL problem also called a contextual bandit problem. In our case, classical PaT methods minimize the time to throw the object while constraining it to fall in a desired area relying on a classical ballistic motion and taking into account the physical properties of the robot\citep{hassan_2022}. However, this approach strongly relies on the properties of the object, the position of the object in the gripper, the accuracy of the sensors and actuators used to measure and control the position of the object, etc. Therefore, as there is uncertainty about these parameters in our model, the trajectory might not always be accurate. Moreover, many of the works on PaT with delta robots use a vacuum gripper which is well-suited for handling smooth, non-porous objects, whereas in our case parallel grippers have been used to handle irregularly shaped and rough objects, such as scrap metal\citep{hassan_2022, raptopoulos_2020}. The jaws of the gripper introduce friction forces and augment the gripper opening delay. These are difficult to model with analytical models and introduce additional unknown parameters. For these reasons, we aim to explore a learning-based solution based on RL to dynamically adapt to varying object geometries, sizes, and conveyor speeds without having to describe the complex dynamic of the environment and estimate the various parameters. To confirm our hypothesis, we compare our approach to the time-optimal PaT S-curves strategy proposed by \citet{hassan_2022} that we adapt to use a parallel gripper. 

This work follows the sim-to-real framework. First, we introduce a novel simulation environment tailored for acquiring PaT strategies through RL. We leverage RL algorithms to train on this task within the designed environment, achieving an 82\% accuracy with a parallel gripper while enhancing the sorting speed compared to the PaP strategy by 13\%. We employ domain randomization during simulation learning. We randomize the opening delay, objects, and positions of the bin to learn a generalized policy that can deal with the dynamics of throwing with a parallel gripper. Our approach is to our knowledge the first one to propose a throwing strategy for delta robots using a parallel gripper or an RL control policy. We conduct an assessment of RL algorithm performances, comparing them with both a PaP approach and a PaT method, both in simulated and real-world scenarios. It is important to note that we have not fine-tuned our policies in reality for the policy transfer to provide a protocol that does not require any additional sensors to replicate our experiments in another industrial line. The outcomes of our experiments show the advantages of RL over PaP and optimization-based PaT methods commonly employed in industrial applications.

In Section \ref{sec:related_work}, we review related works of throwing with robots. In Section \ref{sec:method}, we describe our approach to the problem. In Section \ref{sec:experiments}, we propose two experiments that compare our method to the PaP and the one presented in \citet{hassan_2022} both in simulation and in reality. Finally, Section \ref{sec:conclusion} concludes this paper including possible future works.

\section{Related work}
\label{sec:related_work}
Several works have tried to replace the PaP process with a PaT approach for delta robots.~\citet{raptopoulos_2020} explore the potential to toss an object with a delta robot equipped with a vacuum gripper.~\citet{hassan_2022} extend this possibility and solves an optimization problem based on ballistic motion to determine the ideal trajectory of the robot to throw an item directly into the desired location. A trajectory-planning method for robotic solid waste handling is explored by~\citet{chen2019} where simulations are proposed to validate the benefits of the PaT approach. However, these analytical techniques are limited by the accuracy of the models used to represent the physical system, and they may not be able to handle the uncertainty and variability that is inherent in real-world applications. RL enables one to overcome these limitations because it is a learning-based approach. RL agents do not rely on a priori models of the physical system, but instead learn to control the robot through interaction with the environment. This allows RL agents to adapt to changes in the environment and learn to handle uncertainty.

These techniques for tossing items with delta robots discard the RL approach. However, deep learning has been used to tackle the task of throwing.~\citet{zeng2019tossingbot} proposed a deep learning architecture coupled to a physical model to infer how to throw objects with a seven-degree-of-freedom robot. The model first acquires the pre-throwing condition based on the grasping, and then uses a physics controller and a multi-layer perceptron (MLP) to handle the varying dynamics and performs the throw. The model was able to achieve 600+ grasps per hour with 85\% throwing accuracy. As stated earlier, throwing holds the potential to facilitate the increase in speed of moving objects and enhance overall productivity. However, to the best of our knowledge, no RL technique solves the problem of throwing irregular-shaped objects with parallel grippers at high speed (around 1 item per second) and high accuracy as needed in the industry. Indeed, it is important to create a throwing motion that minimizes time while maintaining efficiency. In this direction, ~\citet{marlier2019robotic} has exposed a first proof of concept with a machine learning method that enabled an ABB IRB 340 robot to throw objects to buckets with an empirical success rate of 99\%. However, the study was only conducted with regular-shaped stones. Moreover, in this solution, the robot first moves in front of the bin to throw perpendicularly to the conveyor and to face the bin which introduces an extra delay before throwing resulting in a less efficient strategy than the classical PaP one. This work aims to go one step further by addressing the challenges of throwing irregular-shaped objects and throwing them from anywhere on the conveyor and beating the PaP process.

\section{Method}
\label{sec:method}
In this section, we first define the problem and its formulation. Afterwards, we detail our training procedure for our models.
\subsection{Background}
We formalize our reinforcement learning problem as a contextual bandit problem. This choice comes from the fact that a single decision has to be made at each throw, and we consider subsequent throws independent, which is an approximation that may be relaxed in future work. A contextual bandit problem is defined with one context space $\mathcal{C}$, a probability distribution $p_0$ over the contexts, an action space $\mathcal{A}$ and a reward function $\mathcal{R}$. We assume a uniform distribution $p_0(c) = \mathcal{U}(c)$ over the contexts in simulation and we only consider linear movement for the delta robot to simplify the model which reduces the size of the action space. We define a policy as a mapping:
 $\pi \colon \mathcal{C} \xrightarrow{} \Delta( \mathcal{A} ) $ from context to distributions over actions. An optimal policy $\pi^{*}$ is a policy that maximizes the expected reward in any context c (Equation \ref{eq:contextual_bandit}),

\begin{equation}
\label{eq:contextual_bandit}
    \pi^* \in {{\arg \max}_\pi} \ {\mathbb{E}_{a \sim \pi(\cdot|c)}}\left[\mathcal{R}(c,a)\right], \quad {\forall {c} \in \mathcal{C}}
\end{equation}
where $\mathcal{R}$ is the reward function depending on the context $c$ and the action $a$ taken by the agent and drawn from its policy $\pi$.

\subsection{Problem statement}

We define the problem by describing the context space, the action space and the reward function.\\

\textbf{Context Space $\mathcal{C}$}
\begin{center}
    $\mathcal{C} =\{((x_o, y_o), (x_b, y_b))\}\in \mathbb{R}^4$
\end{center}

\begin{figure}
    \centering
    \includegraphics[width=0.4\textwidth]{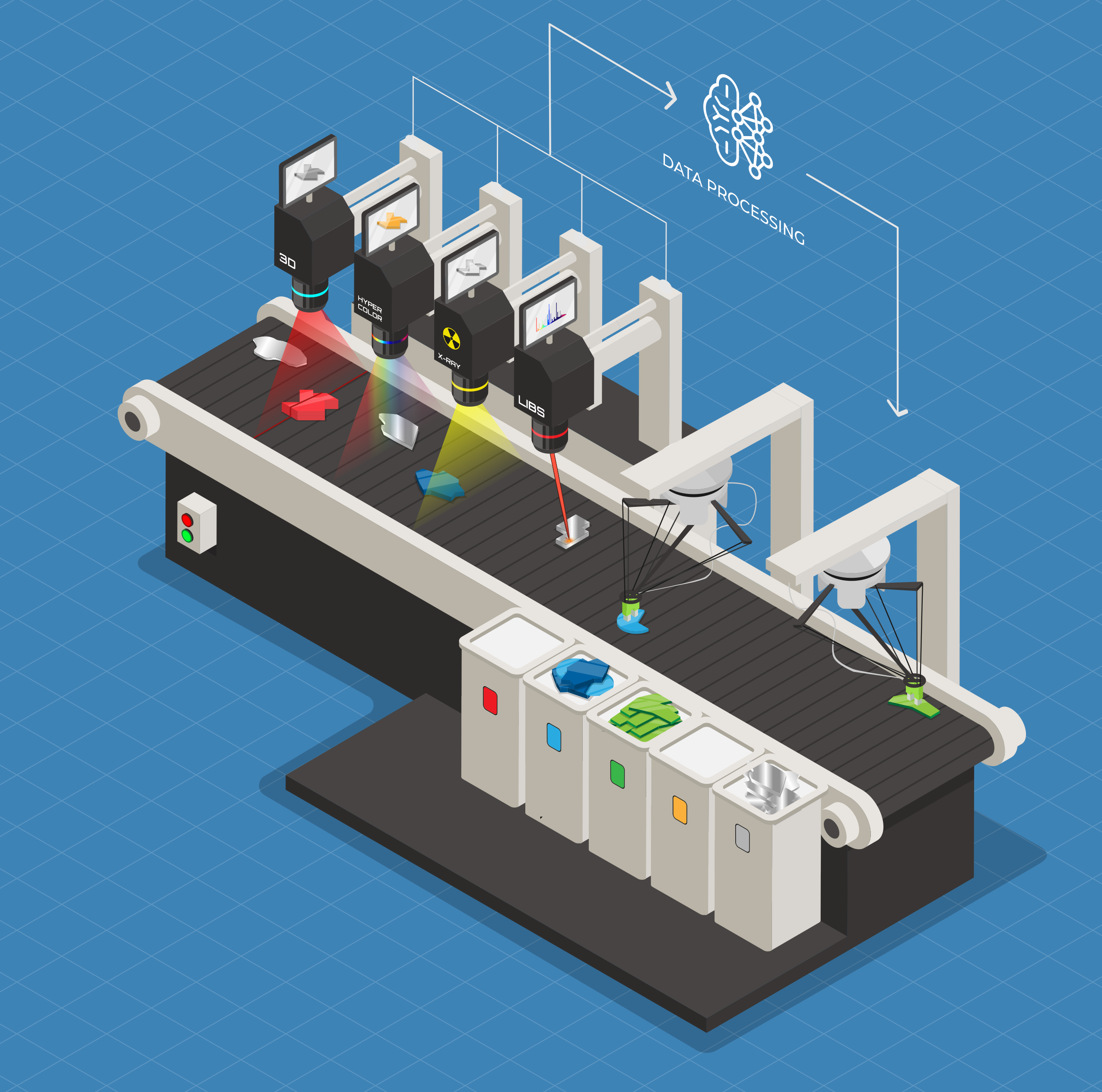}
    \caption{Schema of the PICKIT project with the different sensors (3D, Infrared,  X-ray and LIBS) followed by the delta robots sorting the materials.}
    \label{fig:pickit}
\end{figure}

We denote ($x_o, y_o$) as the position of the object, ($x_b, y_b$) as the position of the bin. These variables define the information available to the agent to select one action. The position of the object is estimated through a 3D camera and the bin is selected with a machine learning model that classifies the object thanks to multiple sensors such as 3D, X-ray transmission and Laser-Induced Breakdown Spectroscopy (LIBS) among others (see Figure \ref{fig:pickit}). \\

\textbf{Action space $\mathcal{A}$}\\

\begin{figure}
    \centering
    \includegraphics[width=0.8\textwidth]{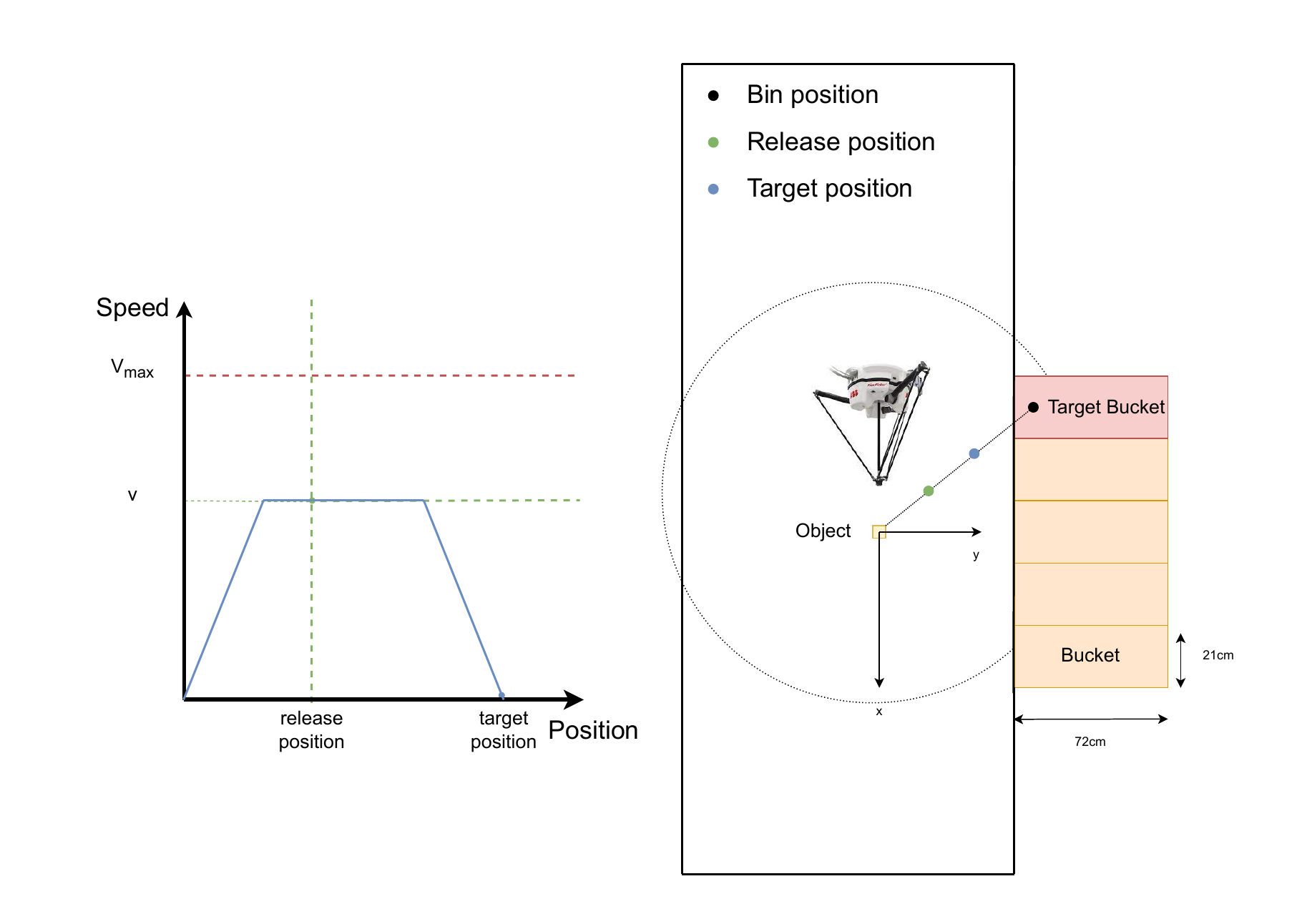}
    \caption{On the left, the speed profile of a throw; On the right, the schema of a throw in the environment.}
    \label{fig:throw_schema}
\end{figure}

When the robot moves it enters first into an acceleration phase before reaching a plateau and decelerates until it arrests its movement. In this work, we suppose a constant acceleration. To throw an object with a given speed a release position is defined where the gripper starts to open before the robot reaches a target position. In addition to the position, we add the velocity the robot has to reach (see Figure \ref{fig:throw_schema}).\\

\begin{center}
    $\mathcal{A} =\{(y_r, v, (z_t, y_t))\}\in \mathbb{R}^4$
\end{center}

We note ($x_t, y_t, z_t$) the target position for the robot after deceleration, $x_t$ is determined by the constraint where the point is in the plane between the object at the pick position and the bin position perpendicular to the conveyor, $v$ is the velocity, while ($x_r$, $y_r$, $z_r$) the position where we start to open the gripper. Note that we only need $y_r$ as the target position since the robot is located on the line between the pick position and the target position. \\

\textbf{Reward function $\mathcal{R}$}\\

The reward function is crucial for solving the problem as it will influence the agent in the learning of its policy~\citep{dewey2014reinforcement, gupta2022unpacking}. The objective is to train the robot to throw objects with high accuracy while operating swiftly. Therefore, we need a fair trade-off between velocity and accuracy in the reward function. Formally, the two quantities we consider are the success of a throw and the total time for the throw. The naive approach weights these two quantities with a parameter. However, setting the value of the parameter is challenging and will greatly influence the final policy. 
Furthermore, to discourage the agent from disregarding challenging tosses, we provide higher rewards for successful tosses in which the object is farther from the bin compared to those in closer proximity to the bin. As this distance is strongly correlated to the time the PaP process takes and this policy is the one we aim to beat, we decided to use the PaP policy as a baseline. The idea was to reward the success of a throw based on the time $b(c)$ the Pick-and-Place process takes to move an object to a bin. This is, therefore, a measure of the difficulty of a toss and additionally, it is expressed in seconds. Moreover, this baseline allow us to directly observe if the agent performs better than the PaP routine.
\begin{center}
    \begin{equation}
    \mathcal{R}(c, a) =
        \begin{cases}
         b(c) - t & \text{if the toss is successful}\\
         - t  & \text{if the toss is unsuccessful}\\
        \end{cases}
    \label{eq:reward}
    \end{equation}
\end{center}

 The reward is expressed in seconds and if the reward is positive, that means that we achieve a better performance than the baseline. However, the PaP policy's time for a given context is not straightforward and directly available. As a consequence, the idea is to build an estimator of the baseline that will be used at training time. In our case, this estimator will be a neural network trained using the time taken by the PaP policy.

\subsection{Training}

\paragraph{Reward baseline.}

An MLP comprised of one hidden layer with 100 hidden units and ReLU activation functions is used as an estimator for the PaP approach. It is trained to minimize the mean squared error with respect to the time taken by the PaP at 10 m/s in simulation. The estimator was afterwards evaluated on 10000 objects with a mean error of 2.5ms.

\paragraph{Hyperparameters optimization}

This work uses Optuna~\citep{optuna_2019}. Optuna is a popular open-source framework for hyperparameter optimization, which can be used to automate the search for the best hyperparameters for a reinforcement learning algorithm. It was used to optimize the hyperparameters of three different reinforcement learning algorithms: SAC~\citep{SAC}, PPO~\citep{PPO}, and TD3~\citep{TD3}. We ran a study of 100 trials with a budget of 50000 episodes for training and 10000 episodes for evaluation. The search space for each algorithm's hyperparameters and the resulting best hyperparameters for each algorithm are reported in Appendices \ref{appendix_search_space} and \ref{appendix_hyperparams} respectively.

\paragraph{Domain randomization}

Domain randomization is used to improve the sim-to-real transfer. It involves adding random variations to the training environment, such as the appearance of objects, the lighting conditions, and the physical properties of the environment. This prevents overfitting the simulated environment and promotes generalisation ~\citep{tobin2017domain}. Instead, the agent learns a more general policy that can be applied to a wider range of environments with the hope that it generalizes well to the real environment. Three key parameters were randomized in this work: the positions of the bins, the gripper opening delay and the object properties.
\newpage
\begin{enumerate}
    \item \textbf{Position of the bins.} At each episode, the position of the bin is randomized in an area of 0.6m by 0.7m:

    \[
    w \sim \mathcal{U}(0, 0.6), \quad
    l \sim \mathcal{U}(0, 0.7).
    \]
    These values have been chosen for the robot to be able to reach the bin in each case and be able to perform a PaP policy, in this instance for an ABB Flexpicker IRB 360 that we have used.
    
    \item \textbf{Opening delay of the gripper.} The gripper in reality has a delay when the open signal is sent. It can be broken down into two factors: the delay before the gripper starts to open and the delay from the beginning of the opening until complete opening. The gripper used on our ABB Flexpicker is a Festo HGPL-14-40-A-B. We have measured these delays and obtained respectively 10ms and 171ms. Therefore, the opening delays are randomized around these values:

    \[ 
    d_1 \sim \mathcal{N}(10, 2), \quad
    d_2 \sim \mathcal{N}(171, 5),
    \]
    where $d_1$ is the delay before the gripper starts to open and $d_2$ is the delay to open the gripper completely.

    \item \textbf{Object.} The size and the mass of the object which is always a cube in simulation is randomized. The mass $m$ is uniformly distributed between 0.01 and 2 kg. The side length $a$ of the cube is uniformly distributed between 3 and 6 cm. Moreover, for the center of mass $c$, since it is a two-dimensional variable it can be shifted vertically and horizontally. The shift follows a uniform distribution in a square region of side $\frac{a}{2}$. We can represent this as two separate variables, $c_x$ and $c_y$, which represent the horizontal and vertical displacement of the center of mass respectively:
    
    \[ 
    m \sim \mathcal{U}(0.01, 0.05), \quad
    a \sim \mathcal{U}(3, 6), \quad
    c_x \sim \mathcal{U}\left(-\frac{a}{4}, \frac{a}{4}\right), \quad
    c_y \sim \mathcal{U}\left(-\frac{a}{4}, \frac{a}{4}\right).
    \]
\end{enumerate}

\begin{figure}
    \centering
    \includegraphics[width=0.8\linewidth]{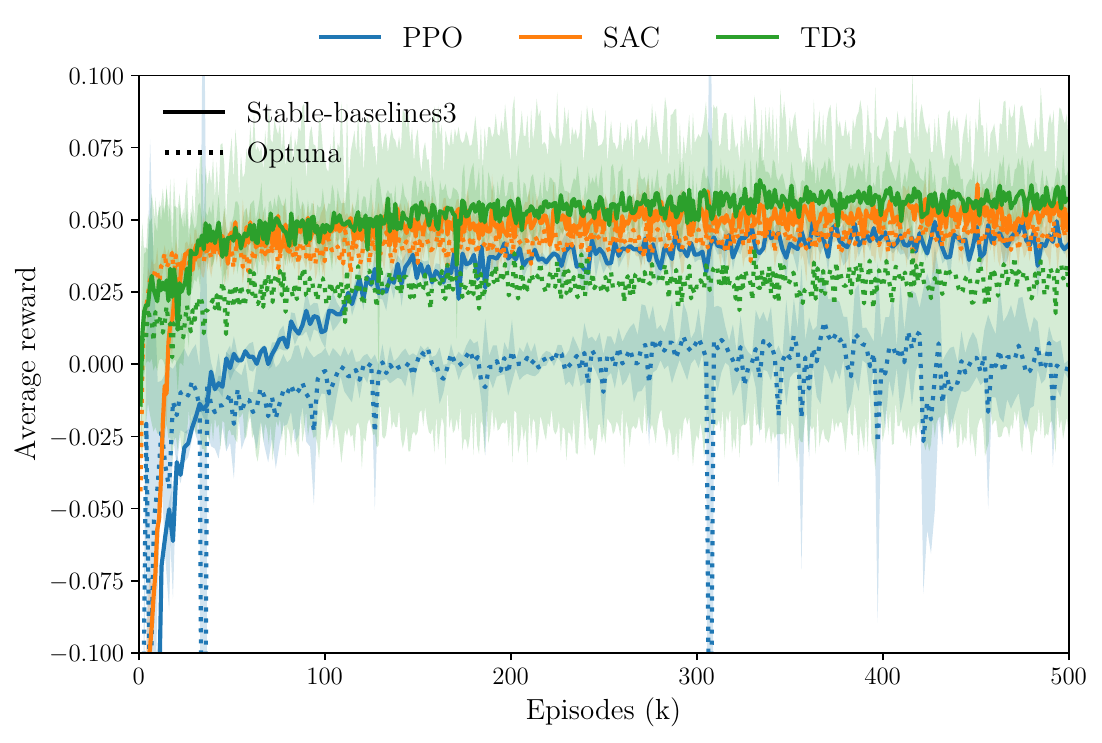}
    \caption{Evolution of the reward for each algorithm (SAC, PPO and TD3) and hyperparameter set (SB3 and Optuna parameters) during the training over 500,000 episodes. For each algorithm/parameter pair we represent the mean between three runs. The intervals represent the standard deviation between the runs.}
    \label{fig:training}
\end{figure}

\paragraph{Model training}
Stable-baselines3 (SB3)~\citep{stable-baselines3} was used for SAC, PPO and TD3. These three algorithms were trained with three different seeds over 500,000 episodes with two sets of hyperparameters: the one obtained with the Optuna study and the default one from SB3. These hyperparameters can be found in the Appendix \ref{appendix_hyperparams}. The learning curves for each algorithm are represented in Figure \ref{fig:training}

\section{Experiments}
\label{sec:experiments}
Two experiments were carried out to evaluate the results of the training. In this section, we start with an experiment in simulation with the corresponding results. Secondly, we present the results obtained in the real-world experiment.

\subsection{Simulation experiment}
First, the policies obtained with the three algorithms with their two sets of hyperparameters (trained over three seeds) were evaluated in simulation. The evaluation metrics used are the success rate to evaluate the accuracy of the throw, the mean distance between the impact of the object and the bin and the distance ratio (the ratio of the distance between the position of the object and the final position of the robot and the position of the object and the position of the bin). The distance ratio aims to see if the agent can throw the object or if its policy converges to a Pick-and-Place policy. The agents were assessed over the randomised environment over 10000 episodes and the results are reported in Table~\ref{tab:exp1_results}. We also compare our results with the PaT strategy using the time-optimal s-curve and the traditional PaP strategy.

\begin{table}[h]
    \centering
    \caption{Evaluation of the performance of each agent in simulation over 10000 episodes with mean score and standard deviation over the randomized environment.}
    \label{tab:exp1_results}
    \begin{tabular}[t]{cccccccc}
    \toprule
    \multicolumn{2}{c}{Algorithm} & Reward (ms) & Success & Time (ms) & Distance ratio & Distance impact (cm)\\
    \midrule
    \multirow{2}{*}{TD3} & SB3 & \textbf{57 $\pm$ 51} & 89.01\% & 107 $\pm$ 39 & 44\% & \textbf{6.9 $\pm$ 27.40} \\
    & Optuna & 29 $\pm$ 65 & 65.08\% & 90 $\pm$ 47 & 34\% & 22.4 $\pm$ 40.6 \\
    \hline
    \multirow{2}{*}{SAC} & SB3 & 54 $\pm$ 56 & 86.41\% & 106 $\pm$ 37 & 43\% & 8.7 $\pm$ 29.1 \\
    & Optuna & 51 $\pm$ 50 & 87.90\% & 113 $\pm$ 43 & 48\% & 8.1 $\pm$ 30.6 \\
    \hline
    \multirow{2}{*}{PPO} & SB3 & 43 $\pm$ 45 & \textbf{89.32\%} & 123 $\pm$ 43 & 57\% & 8.1 $\pm$ 33.1 \\
    & Optuna & -5 $\pm$ 59 & 75.40\% & 146 $\pm$ 64 & 64\% & 14.1 $\pm$ 34.2 \\
    \hline
    Hassan PaT& / & -64 $\pm$ 20 & 0.01\% & \textbf{64 $\pm$ 20} & \textbf{7\%} & 45 $\pm$ 21.8 \\
    \bottomrule
    \end{tabular}
    
\end{table}

First, Table \ref{tab:exp1_results} shows that with a pneumatic gripper even with a small opening delay, the PaT strategy proposed by \citet{hassan_2022} is not able to successfully throw objects into the bin. Indeed, this model is based on a simple ballistic model that ignores the interaction of the object with the jaws of the gripper, and the robot controller inaccuracies. Sometimes the object also collides with the side of the bin. We also observe that even if we remove the delay of the gripper, the model does not capture the complexity of the environment and is therefore unable to succeed.

On the other hand, every RL agent except PPO with the hyperparameters obtained with Optuna beat the PaP strategy by a significant margin. In terms of reward, TD3 with the stable-baseline3 algorithm has been identified as the best algorithm closely followed by SAC. In terms of success percentage, all the algorithms achieve a similar performance around $89\%$ when it converges to an interesting policy. The PaP strategy that goes from the pick position to the bin position takes 180ms on average calculated over 10000 episodes. Our best algorithm gains a significant edge of 73ms on this strategy. We can also see that we achieve a distance ratio of around 40\% for the best strategies compared to the time-optimal PaT proposed by \citet{hassan_2022} which is greedier with only $7\%$. A YouTube demonstration of these policies in simulation is available at \url{https://youtu.be/4I4M09P6n2k}.

\subsection{Real-world experiment}

The simulation aims to build an efficient policy to transfer into a real-life setting. As a consequence, the performances were evaluated in reality with the best agent in terms of scores trained in simulation i.e. TD3 with the default parameters of SB3. It is essential to see that it was achieved without fine-tuning the policies with real data.

We have compared the PaP method with TD3 in real-world conditions. We performed in total 75 throws in total, including three different regions of the conveyor with five different bins and five different pieces. We presume the grasping is successful and report the accuracy and the action time of the robot to throw or place the object. RL PaT achieved a success rate of 82.666\% (62 success out of 75 throws) compared to 100\% for the PaP. Moreover, the PaT average processing time observed was 332 ms, which is 48ms faster compared to the standard Pick-and-Place procedure. As the results demonstrate, the RL PaT strategy transfers well in reality with a small decrease in success rate going down from 89\% to 82\%. More this learned strategy gets an edge of 13\% of cycle time compared to the PaP baseline. We also tested the PaT strategy proposed by \citet{hassan_2022} but it didn't succeed in throwing correctly using a parallel gripper instead of a vacuum one and with more random-shaped pieces. 

A YouTube demo is available at \url{https://youtu.be/Tch0XPKSesI}.

\section{Conclusion}
\label{sec:conclusion}
In this paper, we proposed a novel reinforcement learning approach to enhance the sorting efficiency of scrap metal using delta robots and a PaT process. We explored three different model-free RL algorithms that learn in a simulated environment in order to transfer their policy to delta robots. We developed a novel reward function that is effective for learning the PaT task in simulation. We showed that the RL-based PaT strategy can be transferred to real-world delta robots and achieve decent results without fine-tuning. We used domain randomization in the simulation environment for the transfer in real-life conditions. We compared the RL-based PaT strategy to the conventional PaP strategy and another PaT strategy in reality and simulation. The results show that RL is a promising approach for improving the sorting efficiency of scrap metal using delta robots, even without fine-tuning in a real-world environment.

However, this work also identifies several limitations that should be addressed in future works. Indeed, it was extremely sensitive to the encoded position of the bin center which might be inaccurate in practical settings and does not consider the advantageous shape of some bins. Moreover, the architecture of the bins in this benchmark was not optimized for throwing as they have a very narrow opening to be placed in series. Furthermore, in the formulation of the problem, it would be interesting not to neglect the dependence between successive throws and modelling the problem as a Markov decision process to maximise the cumulative return. This could improve the time gained between two throws by adjusting the trajectories of the delta robot. In addition, since delta robots are capable of executing more complex movements than linear ones, we may extend the capabilities of PaT strategies by considering actions that parametrize more complex movements. Finally, using real-world data to fine-tune in the real-world environment 
could also be a promising path for our approach. We believe that our work is a significant step towards the development of RL-based sorting strategies for delta robots and will inspire other researchers to explore the use of RL for improving the usage of PaT strategies for delta robots in the industry.

\bibliographystyle{plainnat}
\bibliography{references}  
\begin{appendices}
\newpage
\section{Optuna Study}
\label{appendix_search_space}

A tree-structured Parzen estimator (TPE) is used as the optimizer to sample promising regions of the hyperparameter space. In addition, a median pruner is used for to stop trials early on that are unlikely to yield promising results. 
\begin{table}[h]
\centering
\caption{Search Space for TD3 Parameters}
\begin{tabular}{l c}
\hline
\textbf{TD3 Parameter} & \textbf{Search Space} \\
\hline
Learning Rate $\gamma$ & [1e-5, 1e-2] \\
Batch Size & \{16, 32, 64, 100, 128, 256, 512\} \\
Target smoothing coefficient $\tau$ & \{0.001, 0.005, 0.01, 0.02\} \\
Training frequency & \{1, 4, 8, 16, 32, 64\} \\
Noise type & \{ornstein-uhlenbeck, normal, None\}\\
Noise standard deviation & [0,1]\\
Number of hidden units per layer & \{[256, 256], [400, 300]\}\\
\hline
\end{tabular}
\label{tab:TD3_search_space}
\end{table}

\begin{table}[h]
\centering
\caption{Search Space for SAC Parameters}
\begin{tabular}{l c}
\hline
\textbf{SAC Parameter} & \textbf{Search Space} \\
\hline
Learning Rate $\gamma$ & [1e-5, 1e-2] \\
Batch Size & \{16, 32, 64, 100, 128, 256, 512\} \\
Target smoothing coefficient $\tau$ & \{0.001, 0.005, 0.01, 0.02\} \\
Training frequency & \{1, 4, 8, 16, 32, 64\} \\
Learning start & \{0, 100, 500, 1000\}\\
Initial log $\sigma$ & [-4, 1]\\
SDE sample frequency & \{-1, 8, 16, 32, 64\}\\
Number of hidden units per layer & \{[256, 256], [400, 300]\}\\
\hline
\end{tabular}
\label{tab:SAC_search_space}
\end{table}

\begin{table}[h]
\centering
\caption{Search Space for PPO Parameters}
\begin{tabular}{l c}
\hline
\textbf{PPO Parameter} & \textbf{Search Space} \\
\hline
Learning Rate $\gamma$ & [1e-5, 1e-2] \\
Batch Size & \{16, 32, 64, 100, 128, 256, 512\} \\
Entropy coefficient & [1e-9, 0.05]\\
Clip range & \{0.1, 0.2, 0.3, 0.4\}\\
Number of steps per rollout & \{8, 16, 32, 64, 128, \\&256, 512, 1024, 2048\}\\
Number of epochs & \{1, 5, 10, 20\}\\
GAE coefficient $\lambda$ & \{1, 5, 10, 20\}\\
Max gradient norm & \{0.3, 0.5, 0.6, 0.7, 0.8\} \\
Value function coefficient & [0.25, 0.75] \\
Initial log $\sigma$ & [-4, 1]\\
SDE sample frequency & \{End of the rollout, 8, 16, 32, 64\}\\
Number of hidden units per layer & \{[256, 256], [400, 300]\}\\
Activation function & [Relu, Tanh]\\
\hline
\end{tabular}
\label{tab:PPO_search_space}
\end{table}
\newpage
\section{Hyperparameters}
\label{appendix_hyperparams}

\begin{table}[h]
    \centering
    \caption{PPO hyperparameters.}
    \label{tab:ppo_opt}
    \begin{tabular}{ccc}
    \toprule
    Params & Optuna & Stable-baselines3 \\
    \midrule
      Learning rate & 0.0067 & 0.0003\\
      Batch size & 32 & 64\\
      Entropy coefficient & 6.92e-08 & 0\\
      Clip range & 0.4 & 0.2\\
      Number of steps per rollout & 256 & 2048\\
      Number of epochs & 5 & 10\\
      GAE coefficient $\lambda$ & 0.95 & 0.95\\
      Max gradient norm & 0.8 & 0.5\\
      Value function coefficient & 0.49 & 0.5\\
      SDE sample frequency & 16 & only sample at the beginning of the rollout\\
      number of hidden units per layer  & [400, 300] & [64, 64]\\
      Initial log $\sigma$ & -0.52 & 0\\
      Activation function & Tanh & Tanh\\
    \bottomrule
    \end{tabular}
\end{table}
\begin{table}[h]
    \centering
    \caption{TD3 hyperparameters.}
    \label{tab:td3_opt}
    \begin{tabular}{ccc}
    \toprule
    Params & Optuna & Stable-baselines3 \\
    \midrule
    Learning rate & 0.0066 & 0.001\\
    Batch size & 512 & 100\\
    Target smoothing coefficient $\tau$ & 0.02 & 0.005\\
    Training frequency & 8 & 1\\
    Noise type & ornstein-uhlenbeck & None\\
    Noise standard deviation & 0.673 & None\\
    Number of hidden units per layer & [256,256] & [400,300] \\
    \bottomrule
    \end{tabular}
\end{table}
\begin{table}[h]
    \centering
    \caption{SAC hyperparameters from the Optuna study.}
    \label{tab:sac_opt}
    \begin{tabular}{ccc}
    \toprule
    Params &  Optuna & Stable-baselines3 \\
    \midrule
    Learning rate & 0.0016 & 0.0003\\
    Batch size & 16 & 256\\
    Learning starts & 100 & 100 \\
    Training frequency & 4 & 1\\
    Target smoothing coefficient $\tau$ & 0.005 & 0.005\\
    Initial log $\sigma$ & -0.075 & -3\\
    SDE sample frequency & 8 & samples only at the beginning of the rollout\\
    Number of hidden units per layer & [256,256] & [256,256]\\
    \bottomrule
    \end{tabular}
\end{table}
\end{appendices}
\end{document}